\documentclass{article}

\usepackage{PRIMEarxiv}

\usepackage[utf8]{inputenc} 
\usepackage[T1]{fontenc}    
\usepackage{hyperref}       
\usepackage{url}            
\usepackage{booktabs}       
\usepackage{amsfonts}       
\usepackage{nicefrac}       
\usepackage{microtype}      
\usepackage{lipsum}
\usepackage{fancyhdr}       
\usepackage{graphicx}       
\graphicspath{{media/}}     
\usepackage{subcaption}
\usepackage{amsmath}

\begin{document}
 \title{A Large Language Model Pipeline for Breast Cancer Oncology}
\author{
    Tristen Pool \\
    UT Austin \\
    \texttt{tristen.pool@utexas.edu} \\
    \&
    Dennis Trujillo \\
    Mercurial AI Inc. \\
    \texttt{dennis@mercurial-ai.com} \\
}

\maketitle 

\begin{abstract}
Large language models (LLMs) have demonstrated potential in the innovation of many disciplines. However, how they can best be developed for oncology remains underdeveloped. State-of-the-art OpenAI models were fine-tuned on a clinical dataset and clinical guidelines text corpus for two important cancer treatment factors, adjuvant radiation therapy and chemotherapy, using a novel Langchain prompt engineering pipeline. A high accuracy (0.85+) was achieved in the classification of adjuvant radiation therapy and chemotherapy for breast cancer patients. Furthermore, a confidence interval was formed from observational data on the quality of treatment from human oncologists to estimate the proportion of scenarios in which the model must outperform the original oncologist in its treatment prediction to be a better solution overall as 8.2\% to 13.3\%. Due to indeterminacy in the outcomes of cancer treatment decisions, future investigation, potentially a clinical trial, would be required to determine if this threshold was met by the models. Nevertheless, with 85\% of U.S. cancer patients receiving treatment at local community facilities, these kinds of models could play an important part in expanding access to quality care with outcomes that lie, at minimum, close to a human oncologist. 
\end{abstract}

\section{Introduction}

Breast cancer remains one of the most prevalent cancers worldwide, particularly among women, with approximately 1.5 million new cases diagnosed annually. This accounts for 25\% of all female cancer patients \cite{sun_risk_2017}. Access to optimal care for these patients is often limited due to sparse healthcare resources \cite{barrios2022global}. The complexity and variability of cancer treatment decisions necessitate a high level of expertise, which is not always available, particularly in community healthcare settings where 85\% of U.S. cancer patients receive treatment.

The primary problem addressed in this study is the enhancement of breast cancer treatment planning through the use of fine-tuned large language models (LLMs) trained with medical guidelines and a historical patient dataset comprising demographics, genomics, tumor properties, treatments and outcomes. Traditional approaches to oncology require expert-level decision-making, which can be scarce and inconsistent. This study aims to leverage state-of-the-art OpenAI models, fine-tuned with domain-specific clinical datasets and clinical guidelines, to improve the accuracy and consistency of treatment recommendations for adjuvant radiation therapy and chemotherapy.

To address this problem, we employed a novel Langchain prompt engineering pipeline to fine-tune LLMs with specialized oncology data. The training involved a clinical dataset and a corpus of clinical guidelines focusing on two critical factors in breast cancer treatment: adjuvant radiation therapy and chemotherapy. The Duke MRI dataset, consisting of clinical data from 922 breast cancer patients, was used for fine-tuning the models. Key variables such as HER-2 status and tumor stage, which are crucial for treatment planning, were included in the training data.

The outcomes of this study indicate a high classification accuracy (0.85+) for the prediction of adjuvant radiation therapy and chemotherapy. Furthermore, an analysis of the confidence interval for treatment quality compared to human oncologists suggests that the model could potentially outperform human decision-making in 8.2\% to 13.3\% of scenarios. These findings highlight the potential of LLMs to assist oncologists in making more informed and consistent treatment decisions, thereby expanding access to quality care and improving patient outcomes.

By automating components of the cancer care pipeline, this approach not only aims to reduce costs but also to enhance the breadth of treatment considerations beyond the capacity of individual oncologists. This study provides a foundation for future investigations and potential clinical trials to validate the effectiveness of LLMs in real-world oncology settings.

\section{Datasets}
Two main datasets were used within the context of this study, a textualized version of the Duke MRI dataset containing patient genomic, demographic, treatment, and disease progression information and a compendium of select clinical guidelines comprising ASCO and NCCN guidelines for breast cancer. 

\subsection{Duke MRI}
The tabular clinical data contents of the Duke MRI dataset was utilized for finetuning a base GPT 3.5 model with the goal of accurately predicting optimal treatment recommendations for breast cancer patients. The Duke MRI Dataset comprises tabular clinical data representing patient genomics, demographics, and disease progression information along with pre-operative dynamic-contrast-enhanced MRI (DCE-MRI) imagery of 922 breast cancer patients from Duke Hospital \cite{saha2018machine}. For model training purposes, the dataset was divided into two parts: 80\% for training and 20\% for validation, resulting in a validation set consisting of 181 patients.

The clinical data within the Duke MRI dataset includes detailed patient demographics, tumor characteristics, basic MRI findings, and specific treatment information, such as surgery, radiation therapy, and chemotherapy. Additionally, the dataset contains follow-up information, including tumor response and recurrence data. However, since the image data is not relevant for the training of the large language model (LLM), the technical details of the MRI scans were excluded from the clinical dataset used for this purpose.

Two key treatment variables were selected from the Duke MRI dataset to serve as target variables for classification using the fine-tuned LLMs: adjuvant radiation therapy and adjuvant chemotherapy. Both of these variables are binary indicators, specifying whether each patient received the respective treatments. The adjuvant radiation therapy variable indicates whether a patient underwent radiation therapy following the primary treatment. Similarly, the adjuvant chemotherapy variable shows if a patient received chemotherapy as an additional treatment.

Significantly, HER-2 status and tumor stage are critical factors in planning adjuvant treatment for breast cancer, and both of these variables are included in the Duke MRI dataset \cite{Adjuvant_therapy}. These factors play a crucial role in determining the appropriate adjuvant treatments and were therefore integral to the model's classification tasks.

The Duke MRI dataset is particularly well-suited for natural language processing (NLP) tasks due to its rich clinical data, structured format, inclusion of key variables, binary treatment indicators, and follow-up information. The dataset contains extensive clinical information, such as patient demographics, tumor characteristics, and treatment details, which provides a robust foundation for training NLP models to understand and predict treatment outcomes. 

The well-organized and tabulated clinical data aids in efficiently mapping patient information to treatment decisions. The inclusion of crucial variables like HER-2 status and tumor stage, which are significant in breast cancer treatment planning, allows the model to make more accurate predictions by leveraging important clinical factors. Additionally, the binary indicators for adjuvant radiation therapy and chemotherapy simplify the classification task, while the follow-up information, including tumor response and recurrence, provides additional context for the model, enhancing its predictive capabilities. Using a dataset derived from actual clinical settings further ensures that the model is fine-tuned to handle real-world oncology data, making it more applicable and useful in practical healthcare scenarios.

\subsection{Clinical Guideline Corpus}
A custom collection of clinical guidelines was compiled as a starting point to the fine-tuning pipeline, consisting of sources from the American Cancer Society, the American Society of Clinical Oncology (ASCO), and the National Health Service of England, among other reputable institutions. 
The topics of this collection vary from pain management strategies \cite{mao2022integrative} to treatment recommendations
\cite{moy2021chemotherapy}. 
\cite{henry2022biomarkers}, \cite{moy2023chemotherapy}, \cite{burstein2023testing},
\cite{mao2022integrative}, \cite{reck2021five}, \cite{burstein2021endocrine},
\cite{schneider2021management}, \cite{dale2023practical}
In total, before any preprocessing, the corpus consists of approximately 1,048,362 characters. 
The text becomes 742,434 characters after removing sentences with fewer than two nouns. The Q \& A pairs used for training and validation were generated from a subset of this data, according to the Langchain pipeline.
Since oncological recommendations are constantly changing, this corpus is not intended to be static. 
ASCO regularly updates their recommendations, with a new issue being released three times per month. 
\begin{figure*}
\centering
\includegraphics[width=\textwidth]{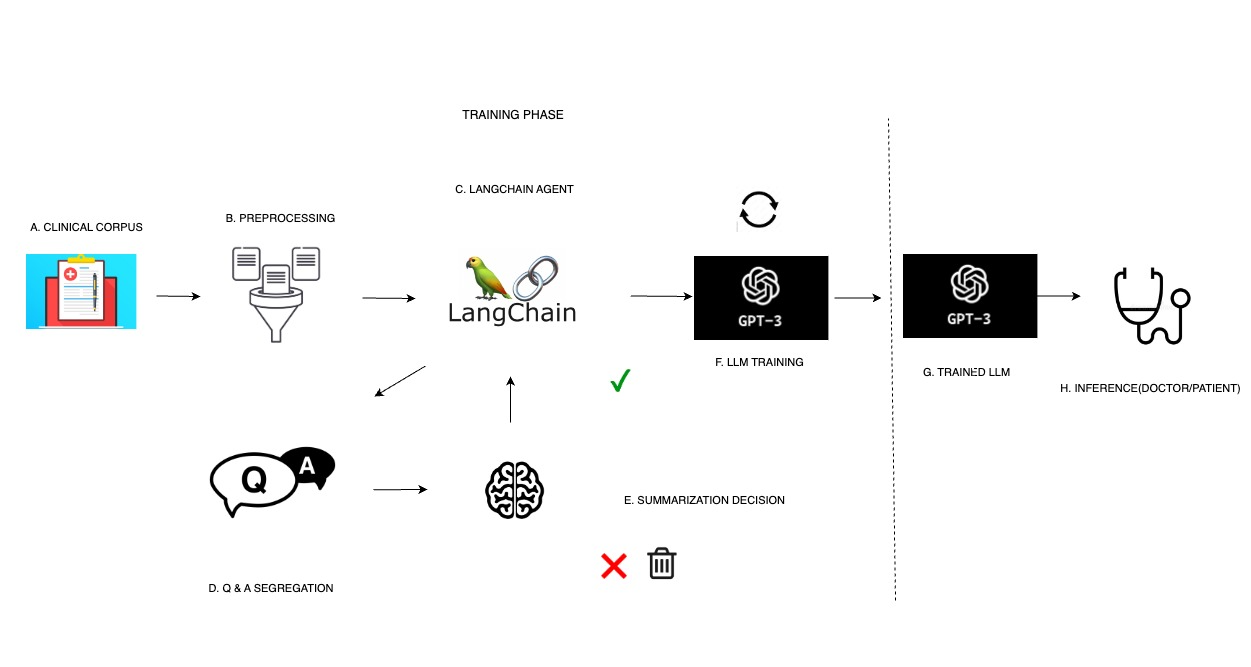}
\caption{The overall architecture diagram. A. Clinical corpus text data, B. Simple text preprocessing, C. Langchain agent handles the decision making, D. The text is then segregated into Q \& A pairs, E. Summarization step where non useful Q \& A pairs are discarded, F. GPT-3 Davinci model trained on the Q \& A pairs, G. The trained model, H. Inference done by doctors/patients/healthcare professionals.}
\label{fig:fig2}
\end{figure*}

\section{Methods}

In this study, we utilized several GPT models developed by OpenAI including GPT-3.5 Turbo, Babbage, and DaVinci which are optimized for interactive chat interactions and dynamic query handling. These base models were chosen for its capabilities in maintaining context across multiple exchanges and providing real-time, contextually relevant responses, which are essential in clinical decision support systems. The implementation of the model involved several key steps, as illustrated in Figure 4.

\subsection{GPT Models}

\subsubsection{GPT-3.5 Turbo}
GPT-3.5-Turbo is designed to excel in scenarios requiring the retention of context through Retrieval Augmented Generation (RAG). This capability is crucial in medical settings where the continuity of patient information and treatment history must be preserved to make informed decisions. The model's ability to provide coherent and relevant responses based on the entire conversation history significantly enhances the quality and reliability of its outputs.

Additionally, GPT-3.5-Turbo possesses advanced function-calling capabilities, allowing it to intelligently select the appropriate tools and parameters for various tasks. In the context of breast cancer treatment planning, this enables the model to adapt its responses dynamically, providing detailed explanations, retrieving relevant clinical guidelines, or generating treatment recommendations as needed. These capabilities make GPT-3.5-Turbo a versatile and powerful tool for interactive clinical decision-making.

\subsubsection{Babbage}

Babbage, also known as ``GPT-3 1B,'' is a 1 billion parameter model developed by OpenAI \cite{brown2020language}. Despite its relatively smaller size compared to other OpenAI models, Babbage is particularly advantageous for classification tasks due to its computational efficiency and cost-effectiveness. The smaller parameter count allows Babbage to perform efficiently without the extensive computational resources required by larger models, making it ideal for tasks that demand lower complexity, such as binary classification of treatment variables. This efficiency is crucial when handling large datasets or when operating under limited computational budgets.

\subsubsection{DaVinci}

Davinci is a 175 billion parameter model developed by OpenAI, known for its superior performance in text completion tasks \cite{brown2020language}. With its extensive parameter count, Davinci excels in generating coherent, contextually relevant, and nuanced text, making it highly effective for detailed language understanding tasks. In the context of this study, Davinci was utilized to generate and refine clinical guideline texts, ensuring that the model could produce high-quality, precise, and contextually appropriate recommendations for breast cancer treatment. Its ability to handle complex language tasks makes Davinci an invaluable tool for developing robust and reliable clinical decision support systems.

\subsection{LangChain}

The methodology for training and utilizing the model involves several key steps. The process begins with compiling a comprehensive clinical corpus, which includes relevant medical texts and clinical guidelines. This corpus serves as the foundational training data. The clinical corpus undergoes preprocessing to ensure the data is clean and structured, involving steps like tokenization and data cleaning.

Following preprocessing, a Langchain agent is employed to manage the decision-making process. This involves using a series of language model chains to generate, verify, and refine question-answer (Q \& A) pairs from the preprocessed data. The generated Q \& A pairs are then segregated, ensuring that each pair is useful for fine-tuning the model, with redundant or irrelevant pairs filtered out.

Subsequently, the Langchain agent makes summarization decisions, determining which Q \& A pairs to retain and which to discard. This step ensures that only high-quality data is used for training. The selected Q \& A pairs are then used to fine-tune the GPT-3.5-Turbo model, adjusting the model’s parameters to enhance its performance on specific tasks related to breast cancer treatment planning.

After training, the fine-tuned model is ready for inference. The model can now provide high-quality, contextually relevant responses to medical queries, assisting doctors and patients. The final step involves using the trained model to support clinical decision-making in real-time, answering medical questions, providing treatment recommendations, and enhancing overall patient care.

\subsection{Duke Pipeline}

The implementation process involved preparing and fine-tuning the model using the Duke MRI dataset, which included detailed clinical data from 922 breast cancer patients. The dataset was divided into training and test sets to ensure robust model evaluation. Texts were tokenized and grouped into blocks to facilitate language modeling. A data collator was used to handle token masking and padding, optimizing the dataset for training.

The training process was configured with specific parameters, such as learning rate, number of epochs, and weight decay, to optimize the model's performance. The training process was facilitated by a robust training framework that enabled efficient handling of the dataset and training loops. The model's performance was continuously evaluated using metrics that measure predictive accuracy, ensuring high-quality outcomes.

GPT-3.5-Turbo was chosen for this study due to its advanced context retention, function-calling capabilities, and optimization for interactive chat-based applications. These features enable it to provide high-quality, contextually relevant, and real-time support for breast cancer treatment planning. Its ability to handle complex medical queries, retain patient-specific context, and dynamically adapt its responses makes it an invaluable tool for enhancing clinical decision-making and improving patient outcomes. By leveraging the strengths of GPT-3.5-Turbo, this study aims to offer a robust and efficient AI-driven solution for breast cancer treatment planning, addressing the critical need for expert-level decision support in oncology.

\subsection{Temperature Sensitivity Analysis}

Temperature is a hyperparameter of LLMs that can drastically alter behavior. 
In text completion tasks the model evaluates the probability that individual words are next in a sequence; these individual probabilities are normalized with a softmax function. 
A lower temperature has a lower effect on the softmax distribution whereas a higher temperature shifts the distribution towards the less likely tokens. 
In order to ensure the models are evaluated at their best, the validation accuracy was generated across the full spectrum of temperature (0 to 2), in iterations of 0.1. The results of this optimization are shown in Figure 1.

Softmax function with temperature:
\begin{equation}
    \text{softmax}(\mathbf{z}, T)_i = \frac{e^{z_i/T}}{\sum_{j=1}^{N} e^{z_j/T}}
\end{equation}

\begin{figure}
    \centering
\includegraphics[width=\textwidth]{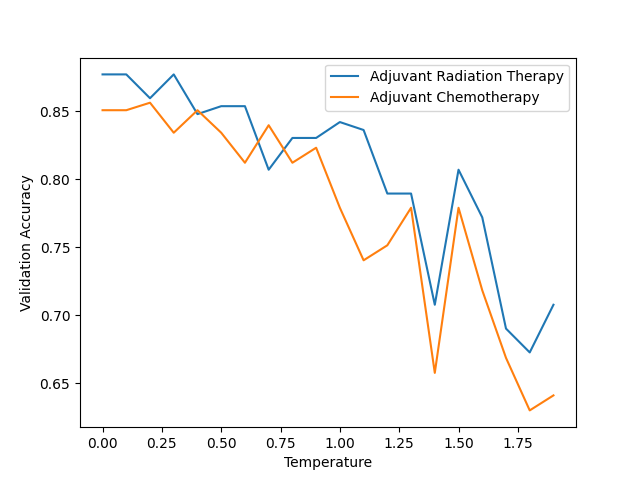}
    \caption{The validation accuracy of the Babbage model fine-tuned for adjuvant radiation therapy classification and adjuvant chemotherapy, both trained for five epochs, plotted against the temperature of the model}
\end{figure}

\begin{figure}[!htb]
    \centering
    \begin{subfigure}[b]{0.51\textwidth}
        \centering
        \includegraphics[width=\textwidth]{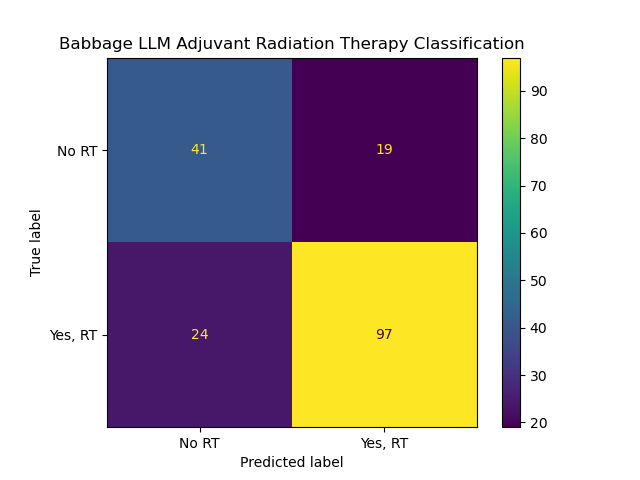}
        \caption{Temperature set to 1 (default)}
        \label{fig:temp_1}
    \end{subfigure}
    \hfill
    \begin{subfigure}[b]{0.48\textwidth}
        \centering
        \includegraphics[width=\textwidth]{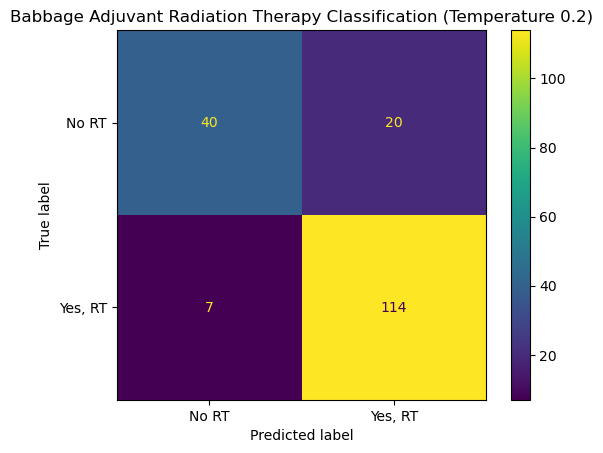}
        \caption{Temperature set to 0.2}
        \label{fig:temp_0.2}
    \end{subfigure}
    \caption{Validation confusion matrices for Babbage adjuvant radiation therapy classification (n=181).}
    \label{fig:confusion_matrices}
\end{figure}

In the case of the adjuvant radiation therapy and chemotherapy models, the fine-tuning data is structured similar to that of a dense neural network, with the clinical features passed in as an array with no textual context. 

This approach makes sense considering the smaller Babbage model is being applied and the intended use-case of the model is with neatly structured tabular data in a clinical database. A comparison between the performance at the default temperature setting and at the optimized temperature for the two treatment factors studied is presented in Figures 2 and 3 with metrics corresponding to the accuracy optimized temperature in Table 1. 

 As the temperature increases from 0.0 to 1.75, the validation accuracy for both models generally decreases, indicating a reduction in model performance. Specifically, the adjuvant radiation therapy model starts with higher accuracy and shows a more gradual decline compared to the adjuvant chemotherapy model, which exhibits more volatility and a steeper drop in accuracy, especially beyond a temperature of 1.25. This suggests that higher temperatures adversely affect the model's ability to correctly classify the therapies, with the chemotherapy model being more sensitive to temperature changes.
\begin{figure}[!htb]
    \centering
    \begin{subfigure}[b]{0.48\columnwidth}
        \centering
        \includegraphics[width=\textwidth]{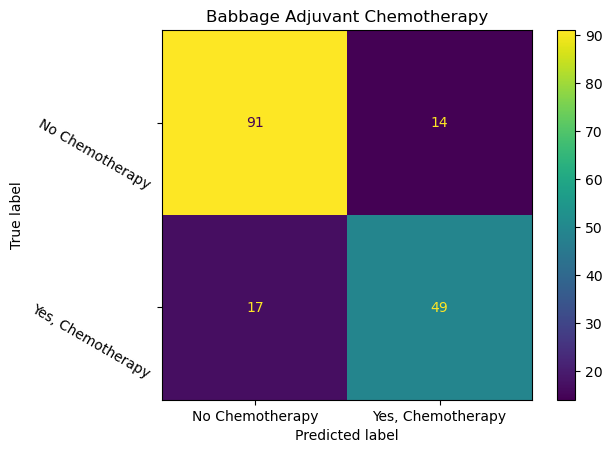}
        \caption{Temperature set to 1}
        \label{fig:chemo_temp_1}
    \end{subfigure}
    \hfill
    \begin{subfigure}[b]{0.48\columnwidth}
        \centering
        \includegraphics[width=\textwidth]{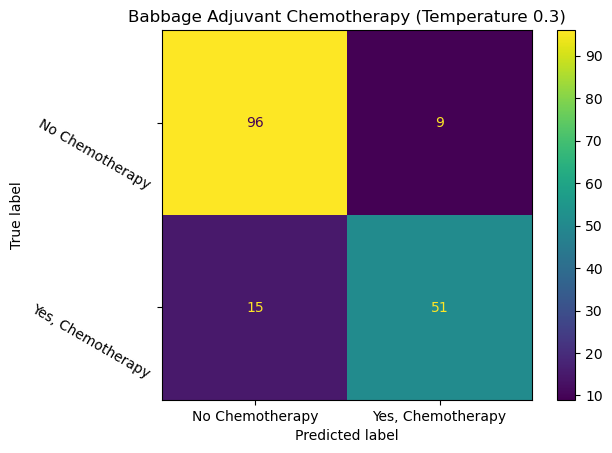}
        \caption{Temperature set to 0.3}
        \label{fig:chemo_temp_0.3}
    \end{subfigure}
    \caption{Validation confusion matrices for Babbage adjuvant chemotherapy classification (n=171)}
    \label{fig:chemo_confusion_matrices}
\end{figure}

\subsection{Clinical Guidelines Pipeline}

Initially, each sentence from the clinical guideline corpus is passed through an array of preprocessing steps including removing whitespace, accented characters, and contractions, as well as filtering out sentences containing a single noun. 
The goal of this process is to help GPT-3.5 in generating the question-answer pairs by simplifying the text data and filtering out sentences that could lead to redundancy. 

Also, removing unnecessary tokens makes the use of the OpenAI API less computationally expensive and more cost-effective.  

Langchain allows LLMs to work in sequence and with other computing tools \cite{Chase_LangChain_2022}. 

As shown in Figure 4, three LLM Chains are employed sequentially, first with a single LLM Chain, then by passing its output to a Simple Sequential Chain, where the output of the other two chains is passed directly as the input to the next. 

The chains are separated from each other in this way only to store the completion from the generative chain. 

The first chain in the sequence converts each sentence into a question and an answer. 

The second verifies that each pair is useful for the purpose of fine-tuning a LLM, eliminating overly general Q\&As or those that could be unrelated to oncology altogether. 

The OpenAI API provides a tool that automates the preprocessing of Q\&A data. 

After these preprocessing steps are applied, the output of the tool in \emph{jsonl} format is partitioned into train and validation sets and used to fine-tune the Davinci model.

A subset of 4500 Q\&A pairs in the total of train and validation sets was taken for purposes of efficiency.

\begin{figure}
    \begin{subfigure}[b]{0.5\columnwidth}
        \includegraphics[width=\columnwidth]{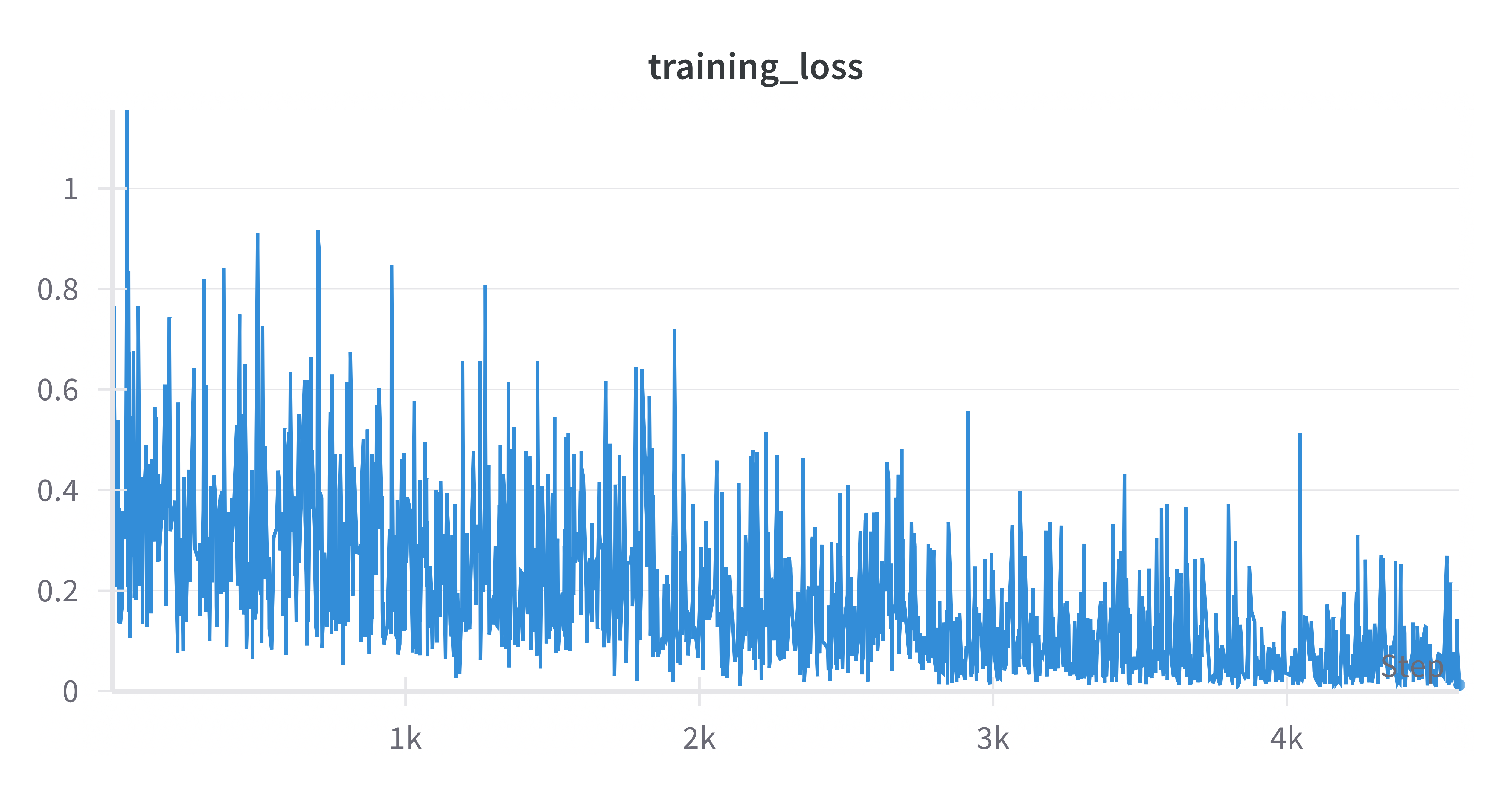}
    \end{subfigure}
    \begin{subfigure}[b]{0.5\columnwidth}
        \includegraphics[width=\columnwidth]{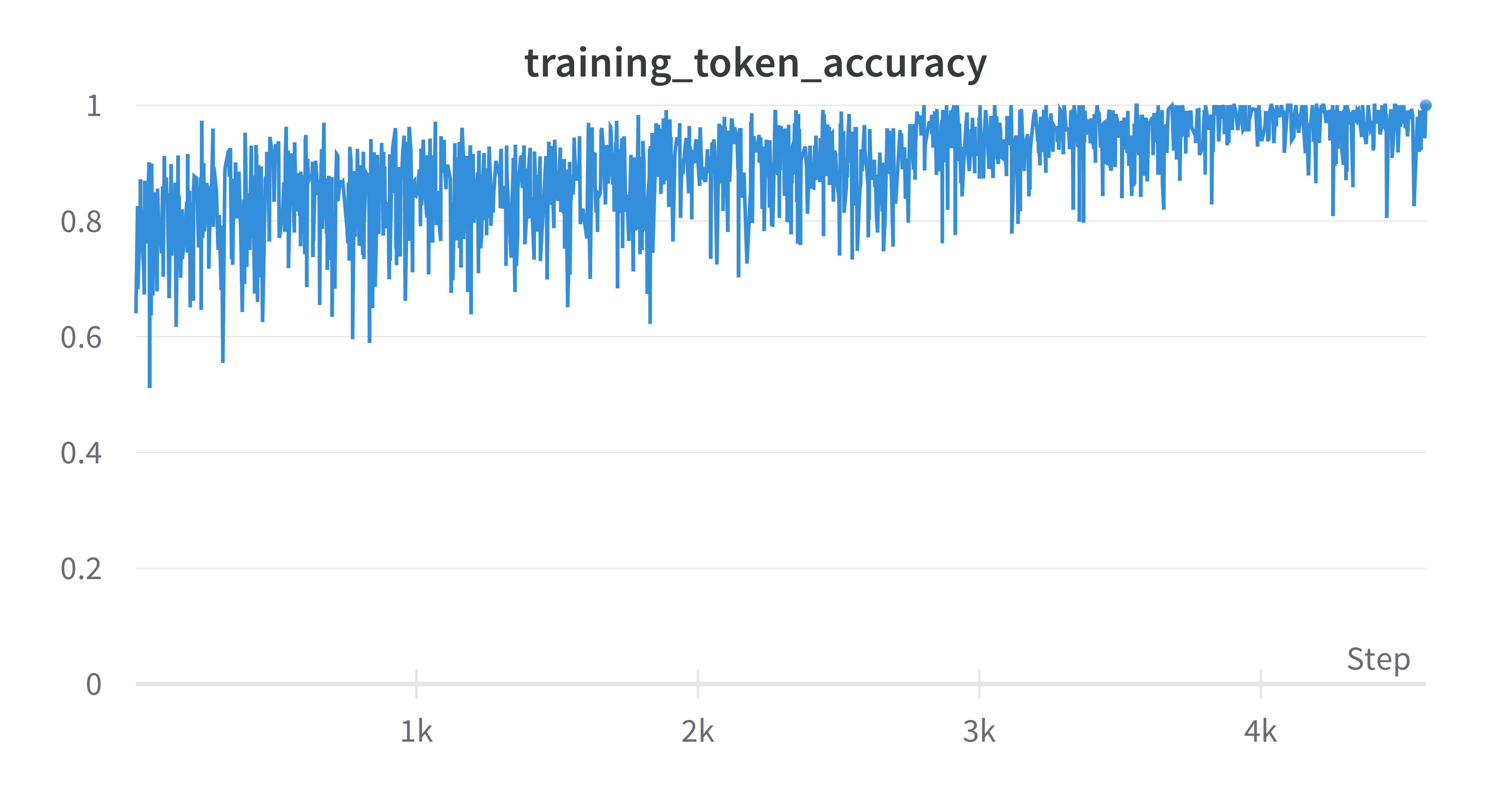}
    \end{subfigure}
    \caption{Davinci chat model training loss (left) and training token accuracy (right)}
    \label{fig:fig3}
\end{figure}

\subsection{Error Analysis}
Since the Duke MRI dataset was entirely formulated with the decisions of human oncologists, the performance data presented in this paper must account for the extent of human error that occurs in the treatment process. A 2019 survey of oncology outpatients found that 13\% believed an error occurred in their care \cite{carey2019patient}. Although this provides a basis, it may not perfectly reflect the specific error rate in the Duke MRI dataset. Ideally, further validation of this error rate with dataset-specific data would strengthen our analysis.

To obtain a confidence interval of the dataset accuracy, assuming that 13\% of cancer treatments involve error, a Wilson score 95\% confidence interval can be calculated:
\begin{equation}
\text{Wilson score interval} = \frac{\hat{p} + \frac{z^2}{2n} \pm z \sqrt{\frac{\hat{p}(1 - \hat{p})}{n} + \frac{z^2}{4n^2}}}{1 + \frac{z^2}{n}}
\end{equation}
where $\hat{p}$ is the assumed sample proportion (0.87), $n$ is the dataset size (922), and $z$ is the z-score corresponding to the 95\% confidence level (1.96). Using this method, it can be concluded with 95\% confidence that the true accuracy of the Duke MRI dataset lies between 0.847 and 0.890.

To enhance the robustness of our error analysis, we also considered alternative methods for calculating confidence intervals, such as the normal approximation interval, and found consistent results.

To calculate the lower-bound and upper-bound of true model accuracy for the prediction of a given variable:
\begin{equation}
\text{Adjusted accuracy} = \frac{TP \cdot (1 - p) + TN \cdot (1 - p)}{n}
\end{equation}
where $TP$ represents true positives, $TN$ represents true negatives, and $p$ is the assumed error rate (0.13). This formula adjusts for human error, providing a more realistic measure of the model's accuracy.

Therefore, without considering the possibility for the model to outperform the human oncologist in some cases, and assuming that there exists some "right" and "wrong" answer to every treatment decision, Table \ref{tab:tab2} illustrates the "true" lower and upper bounds for adjuvant chemotherapy and radiation therapy classification accuracy. Based on this adjusted accuracy, we hypothesize that the model might outperform human oncologists in treatment outcomes, with 8.2\% to 13.3\% of predictions across the two variables being correct while the oncologist was incorrect.

Empirical validation through simulations or further studies is necessary to substantiate these hypothetical improvements in model performance. Moreover, conducting similar analyses on other datasets or in different clinical settings would help generalize these findings, ensuring that the observed error rates and model performance improvements are not unique to the Duke MRI dataset.

\begin{table}
\centering
\begin{tabular}{|c|c|c|}
    \hline
    \textbf{} & \textbf{Adj. Chemo.} & \textbf{Adj. RT} \\
    \hline
    Model Accuracy Lower & 0.728 & 0.721 \\
    \hline
    Model Accuracy Upper & 0.765 & 0.757 \\
    \hline
    Lower-lower Discrepancy & 0.119 & 0.126 \\
    \hline
    Upper-upper Discrepancy & 0.125 & 0.133 \\
    \hline
    Upper-lower Discrepancy & 0.162 & 0.169 \\
    \hline
    Lower-upper Discrepancy & 0.082 & 0.090 \\
    \hline
\end{tabular}
\caption{Confidence interval data for "true" model accuracy, as well as the discrepancy between model accuracy and predicted dataset (human oncologist) accuracy}
\label{tab:tab2}
\end{table}

\section{Discussion}

Fine-tuning LLMs with clinical practice guidelines can increase performance in the domain of oncology.
A chat agent utilizing these developments could be deployed to assist oncologists and improve the care thousands receive each day.
Currently, many oncologists have more data available than they can fully leverage in their practice. The current systems used by medical professionals can even fail at simple tasks like searching through a patient chart; LLM technology offers a solution to this \cite{sorin2023large}. Machine learning technology could also offer improvements compared to human oncologists in edge-cases such as a rare disease that the human is less familar with.
Although this study has arrived at some intuition of the models either merely replicating potentially inadequate decisions or actually outperforming human oncologists, since every patient is unique and cancer treatment is irreversible, there is no way to \emph{guarantee} that a certain treatment is absolutely optimal. However, these models are performant enough to offer a first approximation for a treatment and its predictions, in collaboration with a human oncologists, could improve patient outcomes. 

\bibliographystyle{unsrt}  
\bibliography{bibliography}

\begin{thebibliography}{10}

\bibitem{sun_risk_2017}
Yi-Sheng Sun, Zhao Zhao, Fang Xu, Hang-Jing Lu, Zhi-Yong Zhu, Wen Shi, Jianmin Jiang, Ping-Ping Yao, and Han-Ping Zhu.
\newblock Risk {Factors} and {Preventions} of {Breast} {Cancer}.
\newblock {\em International Journal of Biological Sciences}, November 2017.

\bibitem{barrios2022global}
Carlos~H Barrios.
\newblock Global challenges in breast cancer detection and treatment.
\newblock {\em The Breast}, 62:S3--S6, 2022.

\bibitem{saha2018machine}
Ashirbani Saha, Michael~R Harowicz, Lars~J Grimm, Connie~E Kim, Sujata~V Ghate, Ruth Walsh, and Maciej~A Mazurowski.
\newblock A machine learning approach to radiogenomics of breast cancer: a study of 922 subjects and 529 dce-mri features.
\newblock {\em British journal of cancer}, 119(4):508--516, 2018.

\bibitem{Adjuvant_therapy}
Erin~V Newton.
\newblock Adjuvant therapy for breast cancer.
\newblock {\em Medscape}, March 2022.

\bibitem{mao2022integrative}
Jun~J Mao, Nofisat Ismaila, Ting Bao, Debra Barton, Eran Ben-Arye, Eric~L Garland, Heather Greenlee, Thomas Leblanc, Richard~T Lee, Ana~Maria Lopez, et~al.
\newblock Integrative medicine for pain management in oncology: society for integrative oncology--asco guideline.
\newblock {\em Journal of Clinical Oncology}, 40(34):3998--4024, 2022.

\bibitem{moy2021chemotherapy}
Beverly Moy, R~Bryan Rumble, Steven~E Come, Nancy~E Davidson, Angelo Di~Leo, Julie~R Gralow, Gabriel~N Hortobagyi, Douglas Yee, Ian~E Smith, Mariana Chavez-MacGregor, et~al.
\newblock Chemotherapy and targeted therapy for patients with human epidermal growth factor receptor 2--negative metastatic breast cancer that is either endocrine-pretreated or hormone receptor--negative: Asco guideline update.
\newblock {\em Journal of Clinical Oncology}, 39(35):3938--3958, 2021.

\bibitem{henry2022biomarkers}
N~Lynn Henry, Mark~R Somerfield, Zoneddy Dayao, Anthony Elias, Kevin Kalinsky, Lisa~M McShane, Beverly Moy, Ben~Ho Park, Kelly~M Shanahan, Priyanka Sharma, et~al.
\newblock Biomarkers for systemic therapy in metastatic breast cancer: Asco guideline update.
\newblock {\em Journal of Clinical Oncology}, 40(27):3205--3221, 2022.

\bibitem{moy2023chemotherapy}
Beverly Moy, R~Bryan Rumble, and Lisa~A Carey.
\newblock Chemotherapy and targeted therapy for endocrine-pretreated or hormone receptor--negative metastatic breast cancer: Asco guideline rapid recommendation update.
\newblock {\em Journal of Clinical Oncology}, 41(6):1318--1320, 2023.

\bibitem{burstein2023testing}
Harold~J Burstein, Angela DeMichele, Mark~R Somerfield, and N~Lynn Henry.
\newblock Testing for esr1 mutations to guide therapy for hormone receptor--positive, human epidermal growth factor receptor 2--negative metastatic breast cancer: Asco guideline rapid recommendation update.
\newblock {\em Journal of Clinical Oncology}, 41(18):3423--3425, 2023.

\bibitem{reck2021five}
Martin Reck, Delvys Rodr{\'\i}guez-Abreu, Andrew~G Robinson, Rina Hui, Tibor Cs{\H{o}}szi, Andrea F{\"u}l{\"o}p, Maya Gottfried, Nir Peled, Ali Tafreshi, Sinead Cuffe, et~al.
\newblock Five-year outcomes with pembrolizumab versus chemotherapy for metastatic non--small-cell lung cancer with pd-l1 tumor proportion score $\geq$ 50\%.
\newblock {\em Journal of Clinical Oncology}, 39(21):2339, 2021.

\bibitem{burstein2021endocrine}
Harold~J Burstein, Mark~R Somerfield, Debra~L Barton, Ali Dorris, Lesley~J Fallowfield, Dharamvir Jain, Stephen~RD Johnston, Larissa~A Korde, Jennifer~K Litton, Erin~R Macrae, et~al.
\newblock Endocrine treatment and targeted therapy for hormone receptor--positive, human epidermal growth factor receptor 2--negative metastatic breast cancer: Asco guideline update.
\newblock {\em Journal of clinical oncology}, 39(35):3959, 2021.

\bibitem{schneider2021management}
Bryan~J Schneider, Jarushka Naidoo, Bianca~D Santomasso, Christina Lacchetti, Sherry Adkins, Milan Anadkat, Michael~B Atkins, Kelly~J Brassil, Jeffrey~M Caterino, Ian Chau, et~al.
\newblock Management of immune-related adverse events in patients treated with immune checkpoint inhibitor therapy: Asco guideline update.
\newblock {\em Journal of clinical oncology}, 39(36):4073--4126, 2021.

\bibitem{dale2023practical}
William Dale, Heidi~D Klepin, Grant~R Williams, Shabbir~MH Alibhai, Cristiane Bergerot, Karlynn Brintzenhofeszoc, Judith~O Hopkins, Minaxi~P Jhawer, Vani Katheria, Kah~Poh Loh, et~al.
\newblock Practical assessment and management of vulnerabilities in older patients receiving systemic cancer therapy: Asco guideline update.
\newblock {\em Journal of Clinical Oncology}, 41(26):4293--4312, 2023.

\bibitem{brown2020language}
Tom Brown, Benjamin Mann, Nick Ryder, Melanie Subbiah, Jared~D Kaplan, Prafulla Dhariwal, Arvind Neelakantan, Pranav Shyam, Girish Sastry, Amanda Askell, et~al.
\newblock Language models are few-shot learners.
\newblock {\em Advances in neural information processing systems}, 33:1877--1901, 2020.

\bibitem{Chase_LangChain_2022}
Harrison Chase.
\newblock {LangChain}, October 2022.

\bibitem{carey2019patient}
Mariko Carey, Allison~W Boyes, Jamie Bryant, Heidi Turon, Tara Clinton-McHarg, and Robert Sanson-Fisher.
\newblock The patient perspective on errors in cancer care: results of a cross-sectional survey.
\newblock {\em Journal of patient safety}, 15(4):322, 2019.

\bibitem{sorin2023large}
Vera Sorin, Yiftach Barash, Eli Konen, and Eyal Klang.
\newblock Large language models for oncological applications.
\newblock {\em Journal of Cancer Research and Clinical Oncology}, pages 1--4, 2023.

\end{thebibliography}
  
\end{document}